\documentclass[10pt,twocolumn,letterpaper]{article}

\usepackage{iccv}
\usepackage{times}
\usepackage{epsfig}
\usepackage{graphicx}
\usepackage{amsmath}
\usepackage{amssymb}
\usepackage{bm}
\usepackage{balance}
\usepackage{booktabs}
\usepackage{multirow}
\usepackage{threeparttable}
\usepackage[accsupp]{axessibility}  

\usepackage[pagebackref=true,breaklinks=true,letterpaper=true,colorlinks,bookmarks=false]{hyperref}

\iccvfinalcopy 

\def\iccvPaperID{2909} 

\ificcvfinal\fi

\begin{document}

\title{Explainable Person Re-Identification with Attribute-guided Metric Distillation}

\author{Xiaodong Chen$^1$\thanks{This work was done when Xiaodong Chen was an intern at JD AI Research. }\quad
Xinchen Liu$^2$ \quad
Wu Liu$^2$\thanks{Wu Liu is the corresponding author} \quad
Xiao-Ping Zhang$^{3}$\quad
Yongdong Zhang$^1$ \quad
Tao Mei$^2$\\
$^1$University of Science and Technology of China, Hefei, China \\ 
$^2$JD AI Research, Beijing, China \quad
$^3$Ryerson University, Toronto, Canada\\
{\tt\footnotesize \texttt{cxd1230@mail.ustc.edu.cn, \{liuxinchen1, liuwu1\}@jd.com}, xzhang@ee.ryerson.ca, zyd73@ustc.edu.cn, tmei@live.com}\\
}

\maketitle
\ificcvfinal\thispagestyle{empty}\fi

\begin{abstract}
Despite the great progress of person re-identification (ReID) with the adoption of Convolutional Neural Networks, current ReID models are opaque and only outputs a scalar distance between two persons.
There are few methods providing users semantically understandable explanations for why two persons are the same one or not.
In this paper, we propose a post-hoc method, named Attribute-guided Metric Distillation (AMD), to explain existing ReID models.
This is the first method to explore attributes to answer: 1) what and where the attributes make two persons different, and 2) how much each attribute contributes to the difference.
In AMD, we design a pluggable interpreter network for target models to generate quantitative contributions of attributes and visualize accurate attention maps of the most discriminative attributes.
To achieve this goal, we propose a metric distillation loss by which the interpreter learns to decompose the distance of two persons into components of attributes with knowledge distilled from the target model.
Moreover, we propose an attribute prior loss to make the interpreter generate attribute-guided attention maps and to eliminate biases caused by the imbalanced distribution of attributes.
This loss can guide the interpreter to focus on the exclusive and discriminative attributes rather than the large-area but common attributes of two persons. 
Comprehensive experiments show that the interpreter can generate effective and intuitive explanations for varied models and generalize well under cross-domain settings.
As a by-product, the accuracy of target models can be further improved with our interpreter. \footnote{See the project on~\url{www.xiaodongchen.cn/AMD.github.io/}}
\end{abstract}

\section{Introduction}
\label{sec:intro}
Person Re-identification (ReID), i.e., retrieval of the same person captured by multiple cameras, has attracted tremendous attention from academia and industry~\cite{journals/corr/abs-2104-11536, journals/tip/WangLLLLM20, ijon/WuZZYCZLZJH19, conf/mm/XuHLLS020, journals/corr/abs-2103-12366, iccv/ZhengSTWWT15}.
Although Convolutional Neural Networks (CNNs) have significantly improved the accuracy of person ReID, we still cannot completely trust the results produced by black-box models, especially for critical scenarios~\cite{corr/ZhengYH16}.
Therefore, this paper is focused on the interpretation of CNN-based person ReID models which is crucial yet rarely studied.

\begin{figure}[t]
\begin{center}
\includegraphics[width=0.9999\linewidth]{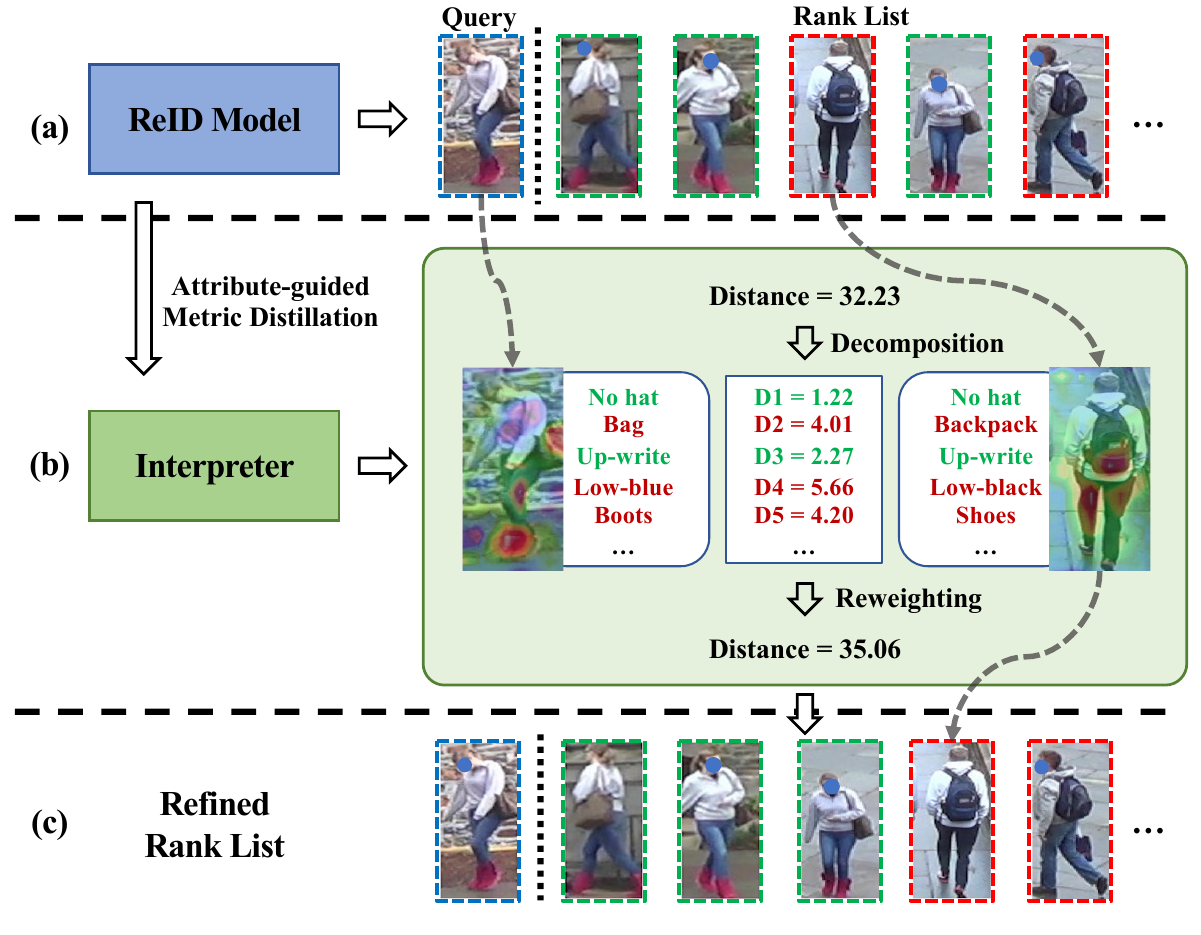}
\end{center}
   \caption{The motivation of attribute-guided metric distillation. (a) Given a query, the ReID model returns a rank list of gallery images based on pairwise metrics. (b) The learned Interpreter can visualize intuitive attention maps of attributes to tell users what attributes make two persons different, and generate contributions of attributes to reflect the impact of each attribute.  (c) Refined results by re-weighted distances from Interpreter. (Best viewed in color.)}
\label{fig:figure1} \vspace{-5mm}
\end{figure}

In recent years, there has been a surge of work in discovering how a target CNN processes input images and makes predictions~\cite{dsaa/GilpinBYBSK18, iccv/SelvarajuCDVPB17, cvpr/ZhouKLOT16}.
These methods usually visualize gradients or salient regions on feature maps w.r.t. the input image and its prediction~\cite{conf/wacv/ChenCHM20, conf/cvpr/DongCH19, iccv/SelvarajuCDVPB17, corr/SimonyanVZ13, conf/wacv/StylianouSP19, cvpr/ZhouKLOT16}.
Particularly, Chen \etal~\cite{iccv/ChenCHRZ19} proposed to explain neural networks semantically and quantitatively by decomposing the prediction made by CNNs into semantic concepts by knowledge distillation.
However, these methods mainly consider classification problems. 
They cannot be directly applied to person ReID, which is an open-set retrieval task and usually solved by metric learning~\cite{iccv/ZhengSTWWT15, iccv/ZhengZY17}.

A CNN-based ReID system usually maps a query image and gallery images into a metric space, then outputs pairwise distances by which a rank list of gallery images is returned, as shown in Figure~\ref{fig:figure1} (a).
Although Yang \etal~\cite{cvpr/YangH0CHZ19} proposed Ranking Activation Maps which could visualize related regions of two persons, it still cannot semantically explain why they are similar or not.
Attributes, e.g., colors and types of clothes, shoes, etc., are semantically understandable for humans and have been exploited as mid-level features for person ReID~\cite{pr/LinZZWHYY19}, but there is no method using attributes for explanations of person ReID.
Therefore, we aim to learn an interpreter with the help of semantic attributes for answering two questions: 1) what attributes make two persons different, and 2) how much impact each attribute contributes to the difference, as shown in Figure~\ref{fig:figure1} (b).
In real applications, the interpreter not only can help users focus on the most discrepant attributes of two persons but also can assist developers to improve the accuracy of ReID models, as shown in Figure~\ref{fig:figure1} (c).

However, interpretation of ReID models with attributes faces unique challenges.
Firstly, since the output of ReID models are distances of pairwise images, it is difficult to use class activation or gradients to visualize salient regions or disentangle semantics as classification~\cite{iccv/SelvarajuCDVPB17, cvpr/ZhouKLOT16}.
Moreover, persons in the wild can be described by various fine-grained and imbalanced attributes~\cite{mm/DENGLLT14, corr/LiZCLH16, pr/LinZZWHYY19}, which may bring biases to ReID results as well as the explanations.
For example, attributes with large areas, such as coats and pants, always overwhelm small but discriminative ones like hats and shoes.
Furthermore, there are only weakly-annotated image-level or ID-level attribute labels without accurate bounding boxes or masks~\cite{pr/LinZZWHYY19}, which makes it hard to learn accurate locations and intuitive visualizations for attributes.

To this end, we propose a post-hoc method, named Attribute-guided Metric Distillation, which explores semantic attributes towards explainable person ReID.
Specifically, we design a pluggable interpreter network to utilize the knowledge from a target ReID model by metric distillation.
The interpreter is grafted on the target ReID model and directly adopts the parameters of the first several CNN stages to exploit the low-level and mid-level features of the target model.
The rest layers of the interpreter are equipped with an attribute decomposition head, by which the interpreter can learn to generate a set of attribute-guided attention maps (AAMs) for a pair of input person images.
On the one hand, the generated AAMs can be directly used to visualize discriminative attributes for the image pair.
On the other hand, the AAMs can be applied to the visual features of the image pair from the target model.
By this means, their features and distance can be decomposed into attribute-guided components to quantify the contribution of each attribute to the overall distance.
Thus, the interpreter not only can output the quantitative contributions of attributes to the overall distance of two persons but also can generate intuitively visualizations of attributes for users.

To guide learning of the interpreter, we design two loss functions.
One is a metric distillation loss which can guarantee the consistency between two distance metrics: 1) the decomposed attribute-guided distances from the interpreter, and 2) the overall distance from the target model.
The other is the attribute prior loss. 
It is designed based on the observation that the difference between two persons mainly comes from the exclusive attributes rather than common ones.
Thus, the attribute prior loss makes the interpreter pay more attention to exclusive but discriminative attributes of two persons with only weakly-labeled attributes of persons.

The contributions of this paper are three-fold: 
\begin{itemize}
\item  This is one of the first attempt toward explainable person ReID by attribute-guided metric distillation that can semantically and quantitatively explain the results of existing ReID models;
\item  We design a pluggable interpreter network with an attribute decomposition head to obtain contributions of attributes to the difference of two persons and generate intuitive visualizations for target ReID models;
\item  To guide the learning of the interpreter, the metric distillation loss and attribute prior loss are proposed to guarantee consistency during metric distillation and prevent biases of attributes.
\end{itemize}
To show the effectiveness and compatibility of the interpreter, we apply it to the state-of-the-art ReID models on different datasets with comprehensive experiments.
As a by-product, the performance of the state-of-the-art models is further improved with our interpreter.

\begin{figure*}
\begin{center}
\includegraphics[width=0.95\linewidth]{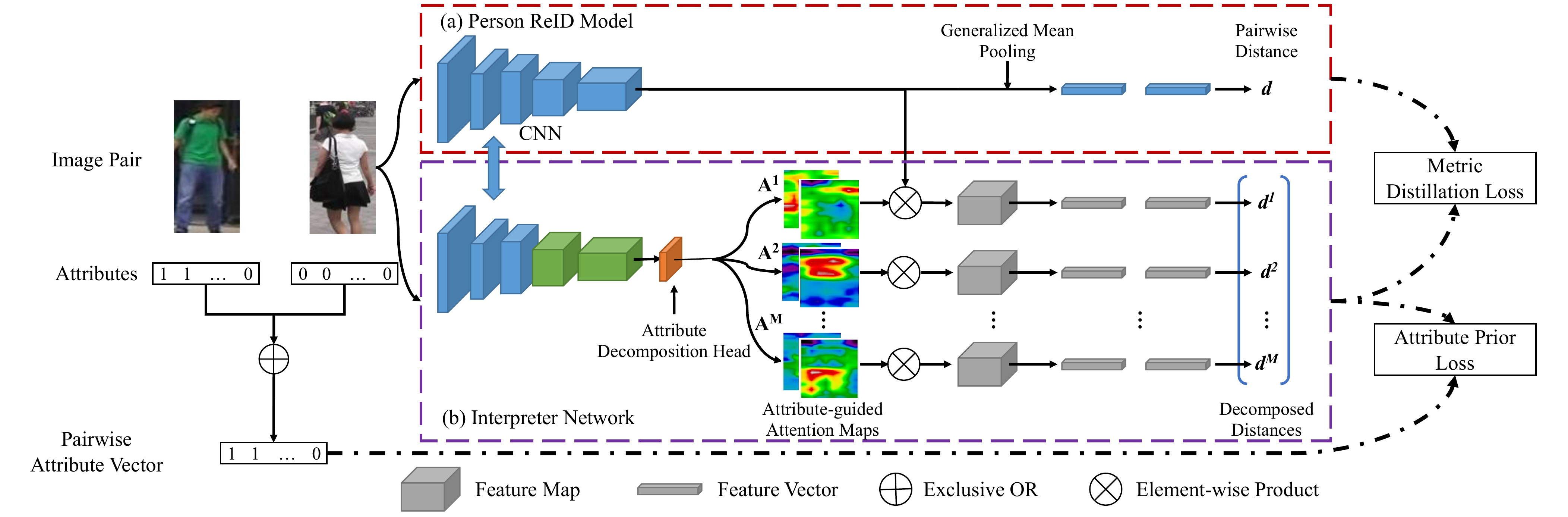}
\end{center}
   \caption{The overall architecture of the attribute-guided metric distillation framework for person ReID. (a) The target ReID model that generates the pairwise distance for an image pair. (b) The interpret network that learns to decompose the pairwise distance into components of attributes and generates attention-guided attention maps for individual attributes. (Best viewed in color.)}
\label{fig:figure2}\vspace{-5mm}
\end{figure*}

\vspace{-0.8mm}
\section{Related Work}
\label{sec:related}
\vspace{-0.8mm}
\textbf{Interpretation of predictions made by CNNs.}
Recent studies on post-hoc methods for the interpretation of CNNs usually adopt visualization of salience maps~\cite{iccv/SelvarajuCDVPB17, corr/SimonyanVZ13, cvpr/ZhouKLOT16}, perturbation of input images~\cite{iccv/FongV17, cvpr/WuSCZKLT20a}, decision trees~\cite{cvpr/ZhangYMW19}, knowledge distillation~\cite{iccv/ChenCHRZ19}, etc.
For example, Simonyan \etal~\cite{corr/SimonyanVZ13} first proposed two techniques based on computing the gradient of the class score with respect to the input.
Zhou \etal~\cite{cvpr/ZhouKLOT16} proposed the class activation mapping (CAM) to map the predicted class score back to the convolutional layers in CNNs, which could provide an intuitive way to highlight the discriminative regions for a specific class.
Zhang \etal~\cite{cvpr/ZhangYMW19} learned a decision tree to clarify the reasons for predictions made by a CNN via estimating the contributions of object parts.
Most recently, Chen \etal~\cite{iccv/ChenCHRZ19} proposed to semantically and quantitatively explain CNNs by knowledge distillation, which can decompose the prediction into contributions of a group of semantic concepts.
Although these methods cannot be directly applied to the interpretation of person ReID models, they inspire us to design a semantic interpreter based on a set of attributes for explainable person ReID.

\textbf{Explainable Person Re-identification.}
Recent CNN-based person ReID methods concentrate on two aspects: 1) task-specific modules to learn discriminative features~\cite{iccv/ChenDH19, iccv/ZhaoLZW17} and 2) metric-based loss functions to make the features of different persons more separable in the latent space~\cite{corr/HermansBL17, cvpr/SunCZZZWW20, ijon/WuZBZYH19, cvpr/ZhengYY00K19}.
Among these methods, various attention modules inspired by the vision system of humans achieve significant performance while making the models more explainable~\cite{iccv/BryanGZP19, cvpr/LiZG18, cvpr/YangH0CHZ19, cvpr/ZhangLZJ020}.
In particular, Yang \etal~\cite{cvpr/YangH0CHZ19} adopted CAM~\cite{cvpr/ZhouKLOT16} to discover rich features for person ReID and proposed Ranking Activation Maps (RAMs) to visualize the salient regions based on the similarity a pair of person images.
However, RAMs only provide rough and pixel-level visualization on images without any semantic explanations.
Therefore, we aims to build a semantic and quantitative interpreter for person ReID models.

\vspace{-0.8mm}
\section{Attribute-guided Metric Distillation}
\label{sec:method}
\vspace{-0.8mm}
\subsection{Preliminary}
\label{subsec:pre}
This subsection first declares necessary notations and definitions for person ReID.
The task of person ReID is, given a query person image, to find images of the same person in a gallery set captured by multiple cameras~\cite{cvpr/LiZXW14,iccv/ZhengSTWWT15}.
CNN-based ReID methods usually follow a paradigm.
\textbf{1)} We have a dataset $S = \{(x_i, y_i)\}^N$ where $x_i$ and $y_i$ are an image and the ID of a person, and $N$ is the number of samples, while $S$ is divided into a training set and a testing set with non-overlapped IDs.
The testing set is further split into a query set $Q$ and a gallery set $G$.
\textbf{2)} In the training stage, a CNN $\mathcal{F}(\cdot)$ is trained to embed the image $x_i$ into a latent space as $\bm{f}_i = \mathcal{F}(x_i)$ by which the features $\bm{f}_i$ of the same person are close and those of different persons are distant.
\textbf{3)} In the testing stage, the trained $\mathcal{F}(\cdot)$ takes a pair of images, $x_q \in Q$ and $x_g \in G$, as the input to obtain their features $(\bm{f}_q, \bm{f}_g)$ and normalized distance $d_{q, g} = \mathcal{D}(\bm{f}_q, \bm{f}_g)$ in the latent space.
In this paper, Euclidean distance is used as the distance metric unless otherwise specified.
By ranking the distances between a query and all images in $G$, the most similar one can be matched.
Through the above paradigm, the prediction of a ReID system is just $d_{q,g}$ for $(x_q, x_g)$.
Given different distance values, \eg $d_{q,g} < d_{q,g'}$, the system cannot semantically and quantitatively explain why $x_q$ and $x_g$ are more similar than $x_q$ and $x_{g'}$, which makes it difficult for users to understand and trust the system.

In this paper, we assume that the person ReID dataset $S$ is weakly labeled with a set of image-level attributes to obtain a new dataset $S_A = \{(x_i, y_i,\bm{a}_i)\}^N$. 
For each image $x_i \in S_A$, $\bm{a}_i = (a^1_i, a^2_i, ..., a^M_i)$ is a binary vector, where $a_i^k$ is the $k$-th attribute denoted by a Boolean value, and $M$ is number of attribute classes.
Our approach aims to semantically and quantitatively explain the distance $d_{i,j}$ with the weakly annotated attributes.

\subsection{Attribute-guided Metric Interpreter}
\label{subsec:interpreter}
This subsection presents design of the interpreter in Attribute-guided Metric Distillation, as shown in Figure~\ref{fig:figure2}.
Before that, we give the formulation of the target model.

\textbf{The target person ReID model} $\mathcal{F}(\cdot)$ can be an arbitrary off-the-shelf model (e.g., PCB~\cite{eccv/SunZYTW18}, MGN~\cite{mm/WangYCLZ18}, BOT~\cite{tmm/LuoJGLLLG20}, SBS~\cite{corr/abs-2006-02631}, etc) with a CNN (e.g. ResNet~\cite{cvpr/HeZRS16}) as the backbone.
$\mathcal{F}(\cdot)$ is trained on the training data as reviewed in Section~\ref{subsec:pre}, then we keep it fixed during learning of the interpreter network.
As shown in Figure~\ref{fig:figure2} (a), given a pair of images ($x_i$, $x_j$), we first extract the feature maps ($\bm{F}_i$, $\bm{F}_j$) from the last convolutional layer.
Then, we obtain the feature vectors ($\bm{f}_i$, $\bm{f}_j$) by generalized mean pooling and compute the distance $d_{i,j}$ in the metric space.

\textbf{The attribute-guided interpreter network} $\mathcal{G}(\cdot)$ is the essential module in the AMD framework.
As shown in Figure~\ref{fig:figure2} (b), the interpreter network has the same structure as the target model.
Since the low-level and mid-level layers of the CNN capture attribute-related features such as texture and color~\cite{cvpr/BauZKO017}, the first several CNN stages of $\mathcal{G}(\cdot)$ and $\mathcal{F}(\cdot)$ are shared to utilize the attribute-related knowledge learned by $\mathcal{F}(\cdot)$.
The high-level layers in $\mathcal{G}(\cdot)$ are learnable to generate the spatial attention maps guided by semantic attributes, which can reflect the contribution of each attribute.

\textbf{An Attribute Decomposition Head} (ADH) is connected after the last convolutional (conv) layer of $\mathcal{G}(\cdot)$.
In particular, the ADH contains a $\frac{C}{8} \times 3 \times 3$ conv layer, a $M \times 1 \times 1$ conv layer, and an activation function $\delta(\cdot)$, where $C$ is the channel number of the last conv layer of $\mathcal{G}(\cdot)$.
Given an image pair ($x_i$, $x_j$), we obtain feature maps from the last conv layer of $\mathcal{G}(\cdot)$.
Through ADH we can obtain the Attribute-guided Attention Maps (AAMs) $\bm{A}_i$ and $\bm{A}_j \in \mathbb{R}^{M \times w \times h}$.
After that, $\bm{A}_i$ and $\bm{A}_j$ are sliced into $M$ matrices by channels, i.e., $(A_i^1,A_i^2,...,A_i^M)$ and $(A_j^1,A_j^2,...,A_j^M)$, where $A_i^k$ and $A_j^k$ $\in R^{h\times w}$ are the attention maps of $k$-th attribute, where $h$ and $w$ are the height and width of the attention maps.

To this end, we can apply $A_i^k$ and $A_j^k$ of attribute $k$ to the feature maps $\bm{F}_i$ and $\bm{F}_j$ from the target model by

\begin{equation}
    \bm{F_i}^k = \bm{F_i} \circ A^k, \quad \bm{F_j}^k = \bm{F_j} \circ A^k,
    \label{equ:equ1}
\end{equation}

where $\circ$ is the element-wise multiplication.
By this means, each input image can obtain $M$ attribute-guided feature maps in which the pixels activated by attribute $k$ will be highlighted, while other pixels will be depressed.
After that, the attribute-guided feature vectors $\bm{f}_i^k$ and $\bm{f}_j^k$ can be calculated from $\bm{F}_i^k$ and $\bm{F}_j^k$ by generalized mean pooling.
Finally, similar to compute $d_{i, j}$ for ($x_i$, $x_j$), we can obtain their attribute-guided distances $(d_{i, j}^1, d_{i, j}^2, ..., d_{i, j}^M)$.

It is noteworthy that the activation function $\delta(\cdot)$ in ADH is important for the generation of AAMs.
For existing attention modules in ReID methods~\cite{cvpr/LiZG18, cvpr/YangH0CHZ19}, the attention maps are usually normalized by a sigmoid activation function to constrain the values in $[0, 1]$.
However, the sigmoid function will make attention values close to either $0$ or $1$, which can cause gradient vanishing and failure of model convergence.
Besides, when multiplying $A^k$ to $\bm{F}$, the large-area parts such as upper clothes and pant will dominant the attribute-guided feature vector $\bm{f}^k$ and make the interpreter learn biased representation for attribute decomposition.
Therefore, we design a Positive Exponential Power Unit (PePU):
\begin{equation}
\delta(x)=  \begin{cases}
    \kappa \cdot (x + 1)^{\tau},   & x > 0, \\
    \kappa \cdot e^x,            & x <= 0,
    \end{cases}
\label{equ:equ2}
\end{equation}
where $x$ is the output of the last conv layer in ADH, $\kappa$ and $\tau$ are growth factor in $(0, 1)$ to smooth attention values and improve propagation of gradients.
With PePU, the interpreter network can effectively decompose the salience regions for various attributes by eliminating biases of imbalanced attribute distribution.

In summary, the interpreter network aims to 1) learn the importance of each attribute for the prediction made by a target model via attribute-guided attention maps, and 2) decompose the distance of two persons into a set of attribute-guided distances based on their contributions to the distance, by which it can provide semantically and quantitatively explanations for person ReID.
To achieve this goal, we elaborately design two types of loss functions in the next section for learning the interpreter by attribute-guided metric distillation.

\subsection{Loss Function}
\label{subsec:loss}

To make the interpreter generate effective and reasonable explanation, we propose two types of objective functions: the metric distillation loss and the attribute prior loss.

\textbf{Loss function of metric distillation.} 
The main task of $\mathcal{G}(\cdot)$ is to decompose the distance $d_{i,j}$ given by the target model $\mathcal{F}(\cdot)$ into the contributions of attributes, which can be formulated as:
\begin{equation}
    d_{i,j} \approx  \hat{d}_{i,j} = \sum\nolimits_{k=1}^{M} d_{i, j}^k,
\label{equ:equ3}
\end{equation}
where $M$ is the number of attributes, $d_{i, j}^k$ is the attribute-guided distance between $x_i$ and $x_j$ for attribute $k$, and $\hat{d}_{i,j}$ is reconstructed distance by the interpreter.
Therefore, we define the metric distillation loss as
\begin{equation}
    L_d = | d_{i, j} - \sum\nolimits_{k=1}^{M} d_{i, j}^k|.
\label{equ:equ4}
\end{equation} 
Different from conventional knowledge distillation for classification, the metric distillation loss can guarantee the consistency between the distance metrics from the target model and the decomposed components from the interpreter.

\textbf{Loss function of attribute prior.}
Only based on the metric distillation loss, the interpreter still cannot decompose the distance in a human understandable way.
As discussed in ~\cite{iccv/ChenCHRZ19}, without any prior knowledge, the explainer tends to suffer from the biased representations, which makes the network tend to approximate the overall distance only by a few dominant attributes instead of discriminative attributes.
To overcome this problem, we define two groups of constraints for the attribute-guided distances. 

The prior constraints are based on the observation that differences of two persons are mainly caused by exclusive attributes like different belongings, rather than common attributes like similar pants.
Therefore, given a pair of input images with attributes, i.e., $(x_i, y_i,\bm{a}_i)$ and $(x_j, y_j,\bm{a}_j)$, the pairwise attribute vector $\bm{a}_{i,j}$ is computed by
\begin{equation}
    \bm{a}_{i,j} = \bm{a}_i \oplus \bm{a}_j,
    \label{equ:equ5}
\end{equation}
where $\oplus$ is Exclusive OR.
With $\bm{a}_{i,j}$, we can obtain the common attributes that both $x_i$ and $x_j$ contain or lack, and the exclusive attributes that only one image contains.

Based on $\bm{a}_{i,j}$, the first group of constraints are applied to the total contribution of exclusive attributes and that of common attributes, which is formulated by:
\begin{equation}
    \sum_{e=1}^{M_E} \frac{d_{i,j}^e}{\hat{d_{i,j}}} \geq  (\frac{M_E}{M})^{\upsilon}, \quad
    \sum_{c=1}^{M-M_E} \frac{d_{i,j}^c}{\hat{d_{i,j}}} \leq 1 - (\frac{M_E}{M})^{\upsilon}.   
    \label{equ:equ6}
\end{equation}
where $\upsilon$ is a factor in $(0,1)$ to regulate the proportion of exclusive attributes, $d_{i,j}^e$ is the distance of an exclusive attribute, $d_{i,j}^c$ is the distance of a comment attribute, $M_E$ is the number of exclusive attributes derived from $\bm{a}_{i,j}$, and $M$ is the number of all attributes.
Through Inequation~\ref{equ:equ6}, the contributions of exclusive attributes tends to be larger than the linear proportion while those of common attributes tends to be smaller.
Based on this prior, we define the first part of the attribute prior loss as:
\begin{equation}
    \begin{aligned}
    L_{p1} & = \max( 0, (\frac{M_E}{M})^{\upsilon} - \sum\nolimits_{e=1}^{M_E} \frac{d_{i,j}^e}{\hat{d}_{i,j}}) \\
           & + \max( 0, \sum\nolimits_{c=1}^{M-M_E} \frac{d_{i,j}^c}{\hat{d}_{i,j}} - 1 +(\frac{M_E}{M})^{\upsilon}).
    \end{aligned}
\label{equ:equ7}
\end{equation}

The second group of constraints is applied to the contribution of individual attribute.
We set a lower bound for each $d_{i,j}^e$ and a upper bound for each $d_{i,j}^c$ formulated by:
\begin{equation}
     \frac{d_{i,j}^e}{\hat{d_{i,j}}} \geq e^{-\lambda} \frac{ ( \frac{M_E}{M})^{\upsilon} }{M_E}, \quad
    \frac{d_{i,j}^c}{\hat{d_{i,j}}} \leq e^{\lambda} \frac{ 1-(\frac{M_E}{M})^{\upsilon}}{M-M_E}. 
\label{equ:equ8}
\end{equation}
Here, we let the above upper bound and lower bound be equal, so the value of $\lambda$ can be solved by 
\begin{equation}
    \lambda = \frac{1}{2} \ln{\frac{M-M_E (\frac{M_E}{M})^{\upsilon} }{M_E (1-(\frac{M_E}{M})^{\upsilon})}}.
\label{equ:equ9}
\end{equation}
From Equation~\ref{equ:equ9}, we can see that the upper bound and lower bound are related to the ratios of exclusive attributes and common attributes to all attributes.
Based on these priors, we define the second part of the attribute prior loss as:
\begin{equation}
\begin{aligned}
    L_{p2} & = \sum\nolimits_{e=1}^{M_E} \max(0, e^{-\lambda} \frac{(\frac{M_E}{M})^{\upsilon}}{M_E} - \frac{d_{ij}^e}{\hat{d}_{ij}})\\
           & + \sum\nolimits_{c=1}^{M-M_E} \max(0,  \frac{d_{ij}^c}{\hat{d}_{ij}} - e^{\lambda} \frac{ 1-(\frac{M_E}{M})^{\upsilon}}{M-M_E}).
\end{aligned}
\label{equ:equ10}
\end{equation}
Through the two attributes prior losses, the interpreter will be more focused on the exclusive attributes that make more contribution to the overall difference of two persons.
Finally, the interpreter is optimized by the total loss function:
\begin{equation}
    L = L_d + \alpha L_{p1} + \beta L_{p2},
\label{equ:equ11}
\end{equation}
where $\alpha$ and $\beta$ are the balance factors.

\subsection{Training and Inference}
\label{subsec:training}
\textbf{Training.}
Firstly, a ReID model $\mathcal{F}(\cdot)$ trained on person ReID data is used as the target model and fixed during learning the interpreter $\mathcal{G}(\cdot)$.
In each training iteration, we take $P \times S$ images as a mini-batch where $P$ is the number of IDs and $S$ is sample number per ID.
In a mini-batch, we can obtain $P^2 \times S^2$ pairs of images to train the interpreter with Equation~\ref{equ:equ11}.
For each pair ($x_i$, $x_j$), we use the distance $d_{i,j}$ generated by $\mathcal{F}(\cdot)$ and the attribute vector $\bm{a}_{i,j}$ as the supervision for training $\mathcal{G}(\cdot)$.

\textbf{Inference.}
During testing, given a query image $x_q$ and a gallery image $x_g$, 
The interpreter $\mathcal{G}(\cdot)$ can generate the attention maps of attributes $(A_q^1,A_q^2,...,A_q^M)$ and $(A_g^1,A_g^2,...,A_g^M)$, and output the attribute-guided distances $\{d_{p,q}^k\}^M$ by forward propagation.
The contribution ratio of attribute $k$ is computed by $r_{p,q}^k = d_{p,q}^k / \sum_{k=1}^{M} d_{p,q}^k$.

\vspace{-0.8mm}
\section{Experiments}
\label{sec:experiments}
\vspace{-0.8mm}
To show the effectiveness and compatibility of the interpreter learned by AMD, we first evaluate our method for different target models on individual datasets.
Then the cross-domain experiments are conducted to demonstrate the generalization of the interpreter.
At last, we incorporate the interpreter with several state-of-the-art ReID models to achieve superior accuracy on different benchmarks.

\subsection{Datasets}
\label{subsec:data}

Our experiments is performed on two large-scale person ReID datasets: \textbf{Market-1501}~\cite{iccv/ZhengSTWWT15} and \textbf{DukeMTMC-ReID}~\cite{iccv/ZhengZY17}.
For convenience, we directly use the ID-level attributes labeled by Lin \etal~\cite{pr/LinZZWHYY19} to train our interpreter.

\begin{figure}[t]
  \centering
  \includegraphics[width=0.9999\linewidth]{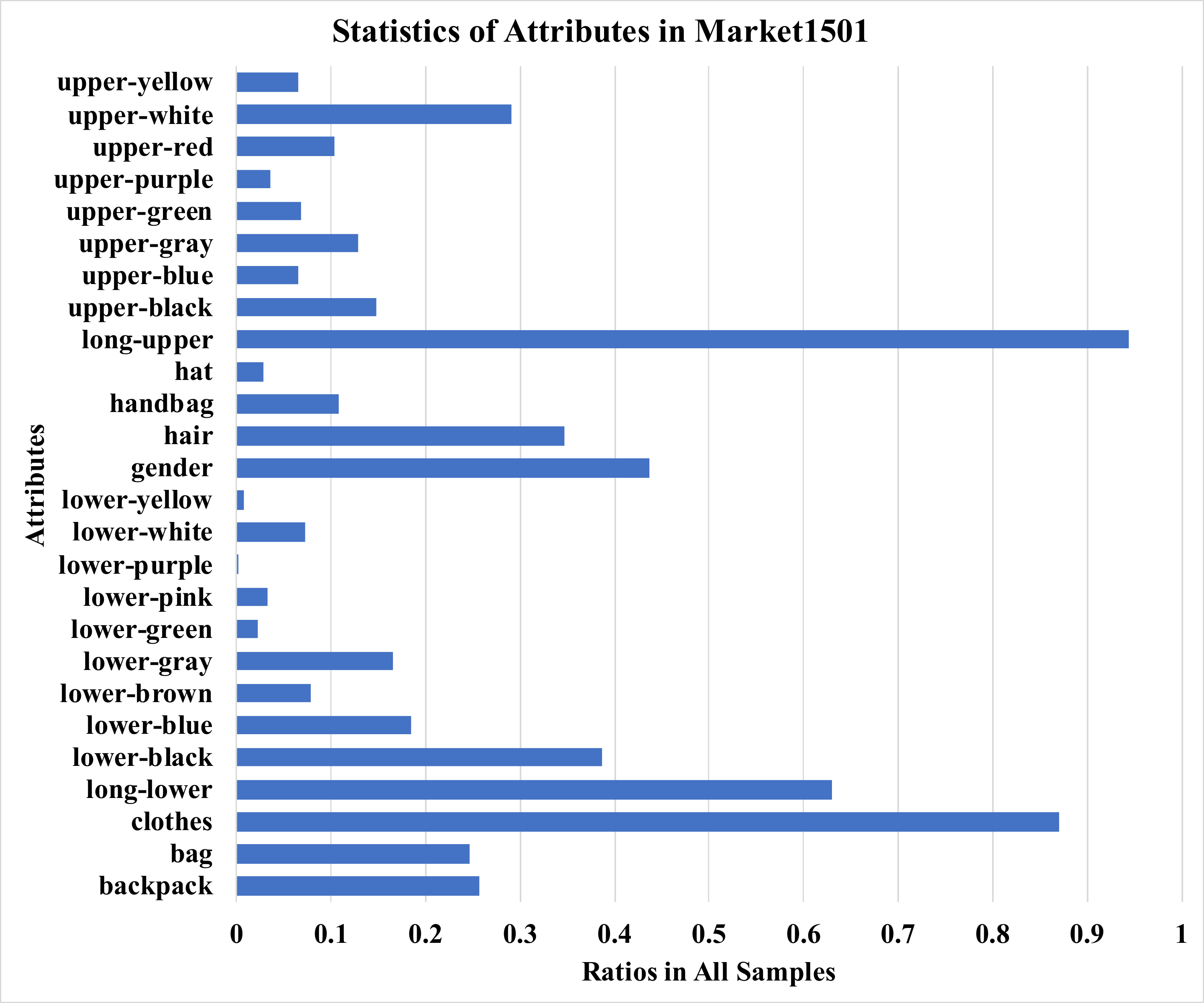}
  \caption{The statistics of attributes on the Market-1501 dataset.}
  \label{fig:figure3} \vspace{-3mm}
\end{figure}

\textbf{Market-1501} contains 751 training IDs with 19,732 images and 750 testing IDs with 13,328 images. 
We select 26 attributes labeled by~\cite{pr/LinZZWHYY19} including: gender (female/male), hair length (long/short), sleeve length (long/short), length of lower clothing (long/short), type of lower clothing (pants/dress), wearing hat (yes/no), carrying backpack (yes/no), carrying handbag (yes/no), carrying other bags (yes/no), 8 colors of upper clothing, and 9 colors of lower clothing.
The statistics of attributes in Market1501 is shown in Figure~\ref{fig:figure3}, which reflects the imbalance of attributes.

\textbf{DukeMTMC-ReID} contains 702 training IDs with 16,522 images and 702 testing IDs with 19,889 images. 
For DukeMTMC-ReID, we select 23 attributes for the interpreter.
For more details on the attributes of DukeMTMC-ReID, please refer to the \textbf{supplementary material}.

\subsection{Implementation Details}
\label{subsec:impl}
This subsection presents the implementation details of our framework and training strategy of the interpreter.

\textbf{Network Structure.}
The ReID model $\mathcal{F}(\cdot)$ and the interpreter $\mathcal{G}(\cdot)$ are built as in Section~\ref{subsec:interpreter}.
For the target models $\mathcal{F}(\cdot)$, we use one of the state-of-the-art ReID models, i.e., Stronger-Baseline (SBS)~\cite{corr/abs-2006-02631}, with different backbones, e.g., ResNet-18/34/50/101.
The interpreter $\mathcal{G}(\cdot)$ uses the same backbone with $\mathcal{F}(\cdot)$ and shares the first three CNN stages, i.e., Conv1 to Conv3, from $\mathcal{F}(\cdot)$.
The rest stages are initialized by parameters pretrained on ImageNet~\cite{cvpr/DengDSLL009}.

\textbf{Networks Training.}
The interpreter $\mathcal{G}(\cdot)$ is trained on the training sets of Market-1501 and DukeMTMC-ReID with the attribute number $M=26$ and $23$, respectively.
We adopt the Adam~\cite{corr/KingmaB14} optimizer to train $\mathcal{G}(\cdot)$ for 30 epochs with the basic learning rate $lr = 10^{-4}$.
The warm-up strategy is used for the first 10 epochs with the initial $lr=10^{-6}$.
For the hyper-parameters in Equation~\ref{equ:equ11}, $\alpha$ and $\beta$ are set to $10.0$ and $50.0$, respectively. 
The $\upsilon$ in Equation~\ref{equ:equ6} is set to $0.5$. 
The $\kappa$ and $\tau$ of PePU in Equation~\ref{equ:equ2} is set to $1/M$ and $0.5$, respectively.
The mini-batch size is $6 \times 4$, \etal, $6$ IDs and $4$ samples per ID.

\subsection{Evaluation Metrics}
\label{subsec:metric}
Although existing interpretation for CNNs are usually demonstrated by visualization and evaluated by subjective observation, we define a group of objective metrics to evaluate the correctness of our interpreter for person ReID.

\textbf{Metric for Distillation.}
Since the interpreter aims to decompose $d_{p,q}$ of ($x_q$, $x_g$) generated by a target model into a set of components, we first measure the information loss during metric distillation from $\mathcal{F}(\cdot)$ to $\mathcal{G}(\cdot)$.
Given the attribute-guided distances $\{d_{q,g}^k\}^M$ from $\mathcal{G}(\cdot)$, we sum all items to obtain a reconstructed distance $\hat{d}_{q,g} = \sum_{k=1}^{M} d_{q, g}^k$.
We report the Average Distance Reconstruction Error (ADRE) over all query-gallery pairs by $\frac{1}{|Q| \cdot |G|}\sum_{q=1}^{|Q|} \sum_{g=1}^{|G|} \frac{| d_{q,g} - \hat{d}_{q,g} |}{d_{q,g}}$.
Moreover, we use the reconstructed $\hat{d}_{q,g}$ to perform the ReID task as using $d_{q,g}$.
Thus, we can observe the information loss by comparing the ReID performance, \eg, Rank-1 accuracy (Rank-1) and mean Average Precision (mAP), of $\mathcal{F}(\cdot)$ and $\mathcal{G}(\cdot)$.

\textbf{Metric for Attribute Decomposition.}
As we expect the interpreter to find the most discrepant attributes of $x_q$ and $x_g$, we measure the ability of $\mathcal{G}(\cdot)$ based on whether it can assign more contributions to the exclusive attributes rather than the common attributes.
In traditional explainability literature, measures such as the pointing game~\cite{conf/eccv/ZhangLBSS16} or insertion/deletion~\cite{journals/corr/abs-2006-03204} are not suitable for our task. 
The ``Point Game'' requires the bounding boxes for attributes, while ReID datasets only have image-level labels.
The ``Insertion/Deletion'' and the ``Blur Integrated Gradients''~\cite{conf/cvpr/XuVS20} are mainly designed for the classification task.
Therefore, we design two metrics X-mAP$_e$ and X-mAP$_c$ for exclusive attributes and common attributes, respectively.

Given input ($x_q$, $x_g$), pairwise attribute vector $\bm{a}_{q,g}$, and the attribute-guided distances $(d_{q,g}^1, d_{q,g}^2, ..., d_{q,g}^M,)$, we rank the distances in an descending order as the larger value means more difference.
Then we compute the Average Precision (AP) of the ranked list like the retrieval task to measure whether exclusive attributes are ranked at the top positions in the list.
The X-mAP$_e$ is the mean value of the AP values over all query and gallery pairs.
Similarly, the X-mAP$_c$ is calculated by ranking the distances in an ascending order to measure whether common attributes are ranked the top positions in the list.
In our experiments, we evaluate the X-mAP$_e$ and X-mAP$_c$ on the testing set for all query and gallery pairs except the pairs from the same ID.


\begin{table*}[t]
\begin{center}
    \footnotesize
    \begin{threeparttable}
    \begin{tabular}{l|c|ccc|ccc}
    \toprule
    Datasets                    & SBS Models    & Rank-1 (\%)   & Rank-5 (\%)   & mAP (\%)  & X-mAP$_e$ (\% $\uparrow$)  & X-mAP$_c$ (\% $\uparrow$) & ADRE (\% $\downarrow$)\\
    \midrule
    \multirow{6}{*}{Market-1501}& ResNet-34     & 93.94 &	97.74 &	83.95    &   -           &   -     & - \\
                                & Interpreter   & 94.21 &	97.80 &	84.12     & 73.71 &	96.39    & 2.31 \\
                                \cmidrule{2-8}
                                & ResNet-50     & 94.77 &	98.13 &	87.15    &   -           &   -     &  - \\
                                & Interpreter   & 94.74 &	98.16 &	87.11    & 74.29 &	96.59   & 1.99 \\
                                \cmidrule{2-8}
                                & ResNet-101    & 95.94 &	98.40 &	88.64     &   -           &   -     &  - \\
                                & Interpreter   & 95.55 &	98.52 &	88.29  & 75.40 & 96.73   & 1.87 \\
    \cmidrule{1-8}
    \multirow{6}{*}{DukeMTMC-reID} & ResNet-34     & 86.67 &	92.77 &	71.71    &   -           &   -     &  - \\
                                & Interpreter   & 86.13 &	92.91 &	72.00   & 69.58 &	95.79  & 1.93  \\
                                \cmidrule{2-8}
                                & ResNet-50     & 88.24 &	94.17 &	75.54  &   -           &   -     & - \\
                                & Interpreter   & 87.84 &	94.34 &	75.27     & 70.30 &	96.03   & 1.73 \\
                                \cmidrule{2-8}
                                & ResNet-101    & 89.33 &	95.20 &	78.41    &   -           &   -     & - \\
                                & Interpreter   & 89.21 &	95.15 &	78.26    & 70.52 &	96.11   & 1.74 \\
    \bottomrule
    \end{tabular}
    \caption{Evaluation of interpreters for different backbone models on Market-1501 and DukeMTMC-ReID. Each target model and the corresponding interpreter are grouped for comparison. The results show that the interpreters learn consistent knowledge to the target models for effective explanations.}
    \label{tab:table1}
    \end{threeparttable} \vspace{-3mm}
\end{center}
\end{table*}

\subsection{Experimental Results of Different Models}
\label{subsec:result}

\begin{figure*}
    \centering
    \includegraphics[width=0.9999\linewidth]{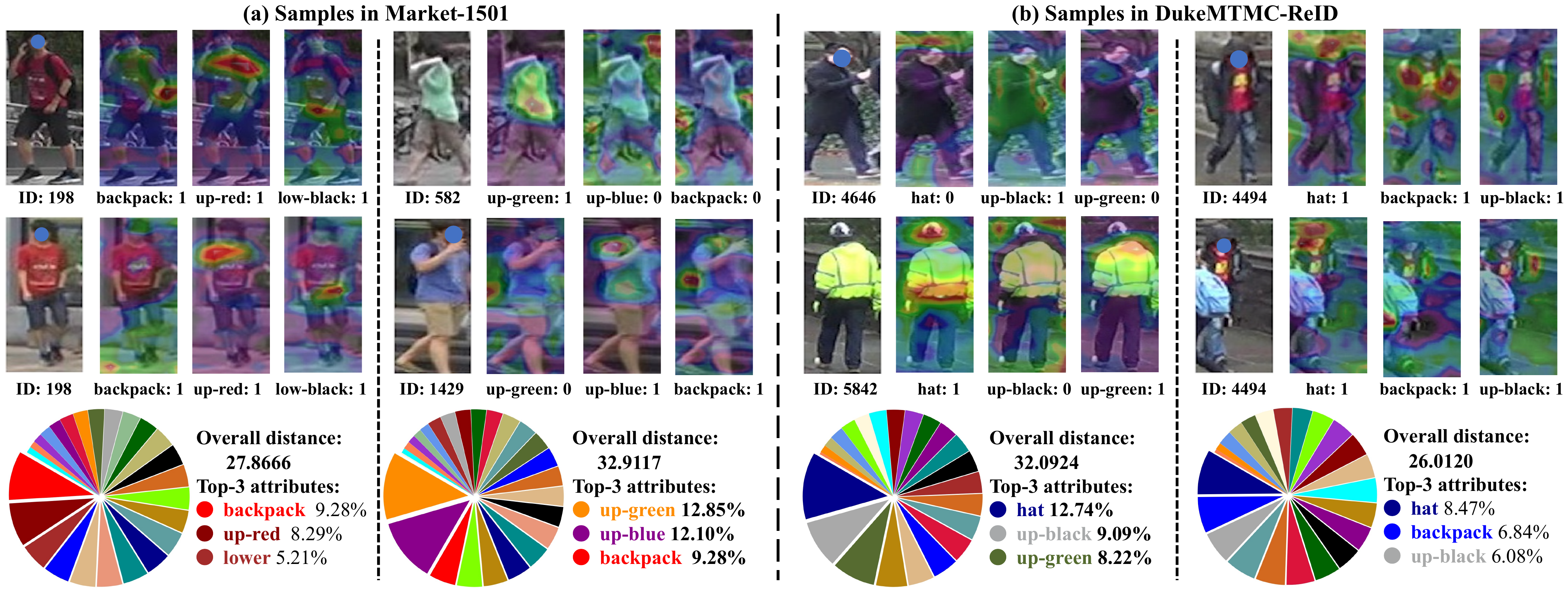}
    \caption{Pairwise examples and explanations for SBS (ResNet-50) on two datasets. For each pair of images, the upper part visualizes the AAMs of the top-3 attributes, which shows that the AAMs are attended to the discriminative attributes. The lower part shows the overall distance and contributions of the top-3 attributes. These figures show the most contributed attributes discovered by the interpreter. (Best viewed in color.)}
\label{fig:figure4} \vspace{-5mm}
\end{figure*}

\begin{table*}[t]
\begin{center}
    \footnotesize
    \begin{threeparttable}
    \begin{tabular}{l|c|ccc|ccc}
    \toprule
    Datasets                    & Models         & Rank-1 (\%)   & Rank-5 (\%)   & mAP (\%)  & X-mAP$_e$ (\% $\uparrow$)  & X-mAP$_c$ (\% $\uparrow$) & ADRE (\% $\downarrow$) \\
    \midrule
    \multirow{2}{*}{M $\rightarrow$ D}
                                & SBS (ResNet-50)     & 45.02 & 61.40         & 26.43     &   -           &   -     &  -  \\
                                & Interpreter   & 47.08 &  62.61        & 28.41     & 59.01        & 91.47  & 2.16 \\
    \cmidrule{1-8}
    \multirow{2}{*}{D $\rightarrow$ M}
                                & SBS (ResNet-50)     & 53.15 & 71.26        & 24.48     &   -           &   -     & -  \\
                                & Interpreter   & 54.48 &  71.59        & 25.48    & 59.19     & 92.17    & 1.56 \\
    \bottomrule
    \end{tabular}
    \caption{Evaluation of the interpreters for SBS (ResNet-50) under the cross-domain setting. M $\rightarrow$ D means the SBS models and interpreters are trained on Market-1501 and tested on DukeMTMC-ReID, and D $\rightarrow$ M means the reverse setting. The results demonstrate that the information loss of interpreters is very minor under the cross-domain setting.}
    \label{tab:table2}
    \end{threeparttable} \vspace{-7mm}
\end{center}
\end{table*}

This subsection presents experiments for interpreting the SBS models with different backbones (ResNet-34/50/101) on Market-1501 and DukeMTMC-ReID.
The metrics including the ReID accuracy of the target model and the interpreter, and X-mAP$_e$, X-mAP$_c$, and ADRE of the interpreter are listed in Table~\ref{tab:table1}.
Examples of attribute-guided attention maps (AAMs) and contributions of top-3 attributes generated for SBS (ResNet-50) are shown in Figure~\ref{fig:figure4}.

In Table~\ref{tab:table1}, the target model and the corresponding interpreter are grouped for comparison.
From ReID accuracy, we can see that interpreters achieve very close accuracy to target ReID models, meanwhile the ADRE between each interpreter and ReID model is also very small.
This means that the information loss is very minor during decomposing the distance from the target model into attribute-guided components by distillation.
The minor loss is acceptable and reasonable because the attribute-guided interpreter sacrifices some discriminative representations but obtains more explainable representations.
Moreover, the good performance of X-mAP$_e$ and X-mAP$_c$ objectively reflects the rationality and correctness of interpreter.
Furthermore, we can find that the metrics of the interpreters are consistent with the target models on different datasets, which reflects the generalization of the interpreter.

In Figure~\ref{fig:figure4}, we show several pairwise images and their explanations for SBS (ResNet-50) models trained on Market-1501 and DukeMTMC-ReID, respectively.
For each pair of images, we visualize the attribute-guided attention maps (AAMs) of the top-3 attributes and list the contributions of top-3 attributes to the overall distance.
From the AAMs, we can observe that for the positive attributes (labeled as $1$), the interpreter can effectively focus on corresponding regions, while for the negative attributes (labeled as $0$), the attentions usually spread around persons.
Particularly, some small objects like backpacks and hats can also be attended to by the interpreter even though they are partially occluded, which shows the effectiveness of the interpreter.

The effectiveness of the interpreter can also be demonstrated by the top-3 contributory attributes.
First of all, the more different the attribute is, the more proportion it contributes to the distance.
Taking the first pair of images as an example, they are the same person captured by different cameras.
Due to varied viewpoints and illuminations, this person looks different especially on the backpack and the color of upper clothes.
The interpreter can effectively assign larger contributions to these discriminative attributes, which can also be observed on other examples.

To further illustrate attentions learned by the interpreter, we show the average attention maps of individual attributes on two datasets in the \textbf{supplementary material}.

\begin{table}[t]
\begin{center}
    \footnotesize
    \begin{threeparttable}
    \begin{tabular}{l|cc|cc}
    \toprule
    \multirow{2}{*}{Models}                 & \multicolumn{2}{c|}{Market-1501}              & \multicolumn{2}{c}{DukeMTMC-ReID}      \\
                                            \cmidrule{2-5}
                                            & Rank-1                & mAP                   & Rank-1                & mAP       \\
    \midrule
    OSNet~\cite{iccv/ZhouYCX19}             & 94.7                  & 85.7                  & 87.9                  & 74.1      \\
     + Re-weighting                         & 95.0\textbf{(+0.3)}   & 86.1\textbf{(+0.4)}                     &      88.5\textbf{(+0.6)}                 & 74.9\textbf{(+0.8)}         \\
    \midrule
    BOT~\cite{tmm/LuoJGLLLG20} (R50)        & 93.8                  & 84.7                  & 86.9                  & 74.3      \\
     + Re-weighting                         & 94.4\textbf{(+0.6)}   & 86.1\textbf{(+1.4)}   & 88.6\textbf{(+1.7)}   & 75.6\textbf{(+1.3)}\\
     \midrule
    CL~\cite{cvpr/SunCZZZWW20} (R50)        & 94.9                  & 85.7                  & 87.1                  & 71.9         \\
     + Re-weighting                         & 95.2 \textbf{(+0.3)}  & 86.4 \textbf{(+0.7)}  & 88.3\textbf{(+1.2)}   & 73.1\textbf{(+1.2)}         \\
    \midrule
    SBS~\cite{corr/abs-2006-02631} (R50)    & 94.8                  & 87.2                  & 88.2                  & 75.5      \\
     + Re-weighting                         & 95.2\textbf{(+0.4)}   & 87.9\textbf{(+0.7)}   & 89.1\textbf{(+0.9)}   & 75.6\textbf{(+0.1)} \\
    \midrule
    SBS~\cite{corr/abs-2006-02631} (R101)   & 95.9                  & 88.6                  & 89.3                  & 78.4      \\
     + Re-weighting                         & 96.1\textbf{(+0.2)}   & 88.8\textbf{(+0.2)}   & 90.2\textbf{(+0.9)}   & 79.1\textbf{(+0.7)} \\
    \bottomrule
    \end{tabular}
    \caption{Comparison between results of the SOTA methods and refined results by re-weighted distances on Market-1501 and DukeMTMC-ReID. For all compared models, the results are further boosted.}
    \label{tab:table3}
    \end{threeparttable} \vspace{-7mm}
\end{center}
\end{table}

\subsection{Cross-domain Evaluation}
\label{subsec:Cross-Domain}
To validate the generalization of our interpreter, we conduct experiments under the cross-domain setting on Market-1501 and DukeMTMC-ReID.
Here we use the SBS (ResNet-50) models trained on Market-1501 and on DukeMTMC-ReID as two target models, then learn two interpreters for these two models.
During the evaluation, we apply the SBS model and the interpreter trained on Market-1501 to the testing set of DukeMTMC-ReID (M $\rightarrow$ D) and vise vase (D $\rightarrow$ M).
The experimental results are listed in Table~\ref{tab:table2}.
From the cross-domain results, we can find that the ReID metrics of the interpreters are still consistent with those of the target models, meanwhile the ADRE values are also very small.
This also reflects that the interpreters can learn consistent knowledge from the target models under the cross-domain setting.
Interestingly, we can find that the results of interpreters are better than those of the target models.
This may be because that the generalization of the attribute-based representations is stronger than the visual features learned by the target models for cross-domain ReID.
Therefore, attributes may have the potential to bridge the domain gap between different datasets.
See the \textbf{supplementary material} for more examples.

\subsection{Distance Re-weighting by Interpreter}
\label{subsec:Reweight}
As a by-product, we try a straightforward method to improve the performance of the state-of-the-art models with the explanations generated by the interpreter.
As in Section~\ref{subsec:training}, our interpreter can output the contributions of exclusive attributes to the overall distance, i.e., $\{d_{p,q}^e\}^{M_E}$ for each pair of query and gallery images.
Then we can simply amplify the component of the most contributed attribute to obtain an updated distance.
Given an original distance $d_{p,q}$, the updated distance can be computed by linear re-weighting as:
\begin{equation}
d_{i,j}^{'} = d_{i,j} + \gamma \cdot \max(\{d_{p,q}^e\}^{M_E}),
\label{equ:equ12}
\end{equation}
where $\gamma$ is a hyper-parameter and set to 1.0 in experiments.

Comparison between results of the state-of-the-art (SOTA) methods and refined results by re-weighted distances from interpreters are listed in Tabled~\ref{tab:table3}.
On both Market-1501 and DukeMTMC-ReID datasets, the performance of all models is improved.
Especially on DukeMTMC-ReID, we obtain 0.9\% and 0.7\% increases in rank-1 and mAP accuracy for the powerful SBS (ResNet-101).
These results demonstrate great potential to explore attributes for further improvement of person ReID.


\section{Conclusion}
\label{sec:conclusion}
This paper presents an Attribute-guided Metric Distillation (AMD) method to use semantic attributes for explainable person ReID.
The AMD learns a pluggable interpreter that can be grafted on any target CNN-based ReID model.
With the metric distillation guided by attribute priors, the learned interpreter can decompose the distance of two person images into quantitative contributions of attributes by which users can know what attributes make two persons different.
Meanwhile, the interpreter can visualize attention maps of discriminative and exclusive attributes to tell users where the most significant attributes are.
With such quantitative explanations and intuitive visualizations, the interpreter can help users make decisions more effectively.
In future work, the proposed AMD framework can be applied for explanation of other metric-based computer vision tasks like content-based image retrieval~\cite{conf/cvpr/DongCH19}, vehicle ReID~\cite{icme/LiuLMF16, tmm/LiuLMM18}, etc.

\textbf{Acknowledgements.} This work was supported by the National Key R\&D Program of China under Grant No. 2020AAA0103800.

\newpage
\begin{appendix}

\section{Supplementary Material}
\label{sec:experiments}

In this supplementary material, we first introduce the attributes of the DukeMTMC-ReID~\cite{iccv/ZhengZY17} dataset.
Then we provide additional experiments for 1) more visualizations of the learned attribute-guided attention maps (AAMs) on Market-1501~\cite{iccv/ZhengSTWWT15} and DukeMTMC-ReID~\cite{iccv/ZhengZY17} datasets, 2) visualizations of AAMs under the cross-domain setting, and 3) analysis on different designs of the interpreter networks.

\subsection{Details of the DukeMTMC-ReID Dataset}
\label{subsec:data}

\begin{figure}[t]
  \centering
  \includegraphics[width=\linewidth]{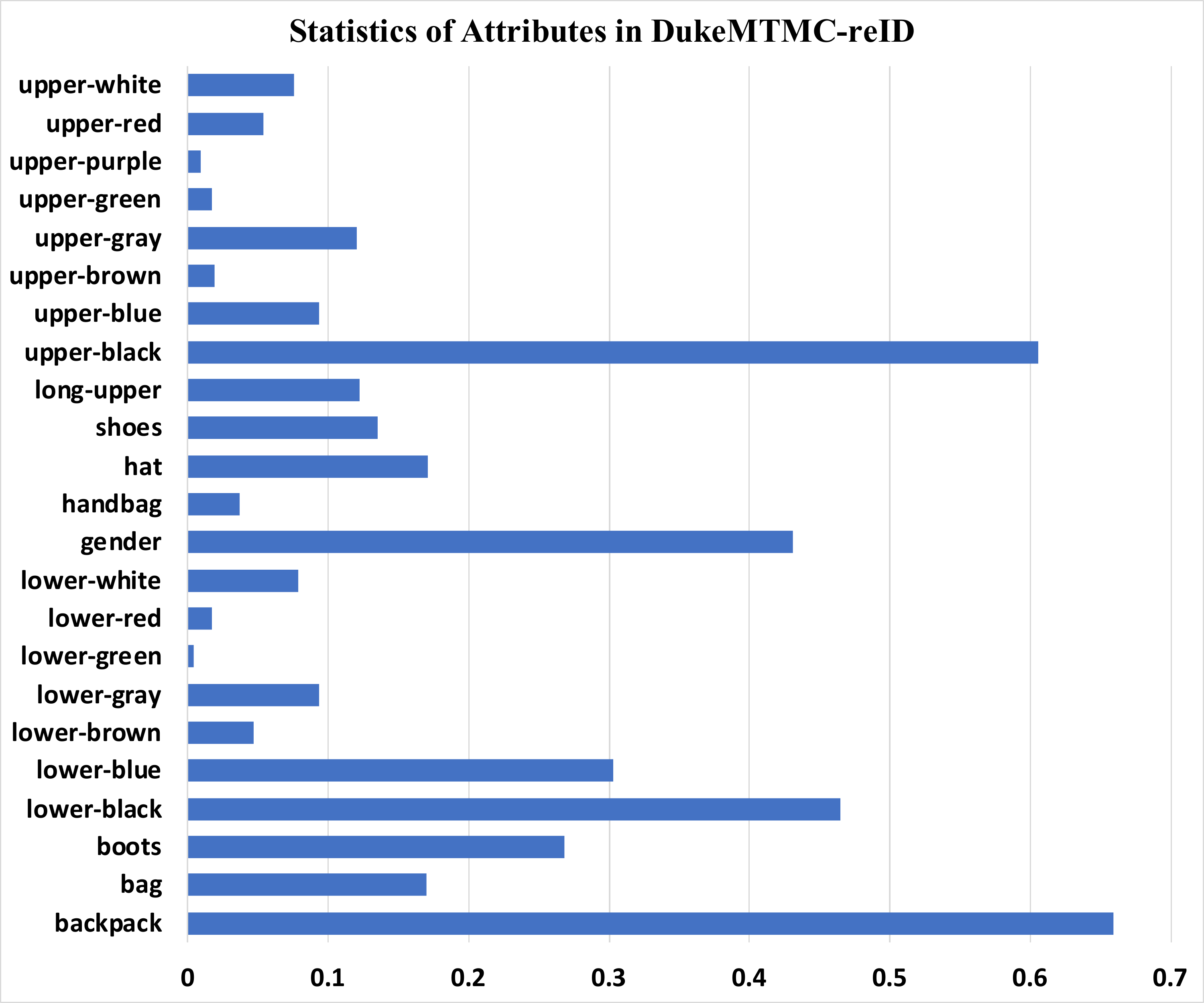}
  \caption{The attribute distribution of DukeMTMC-ReID~\cite{iccv/ZhengZY17}.}
  \label{fig:figure5}
\end{figure}

\begin{figure*}[htbp]
    \centering
    \includegraphics[width=0.92\linewidth]{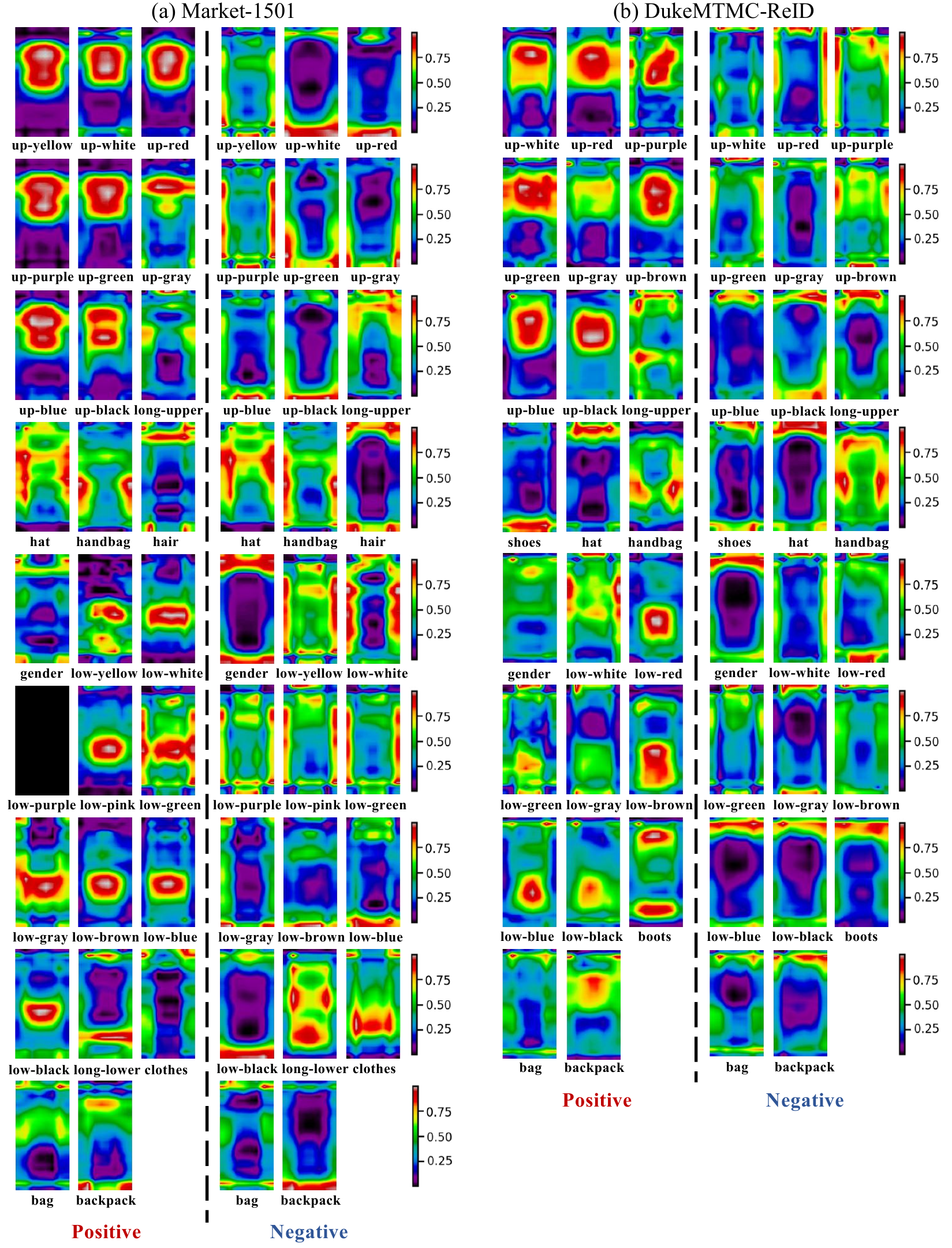}
    \caption{Visualization of average attention maps from the interpreter for the target model, SBS (ResNet-50)~\cite{corr/abs-2006-02631} trained on (a) Market-1501~\cite{iccv/ZhengSTWWT15} and (b) DukeMTMC-ReID~\cite{iccv/ZhengZY17}. In each sub-figure, the left part shows the average attention maps of each attribute which is obtained from all images that contain a certain attribute. The right part shows the average attention maps of each attribute obtained from all images that do not contain that attribute. (Best viewed in color.)}
    \label{fig:figure6}
\end{figure*}

\begin{figure*}[t]
    \centering
    \includegraphics[width=0.9999\linewidth]{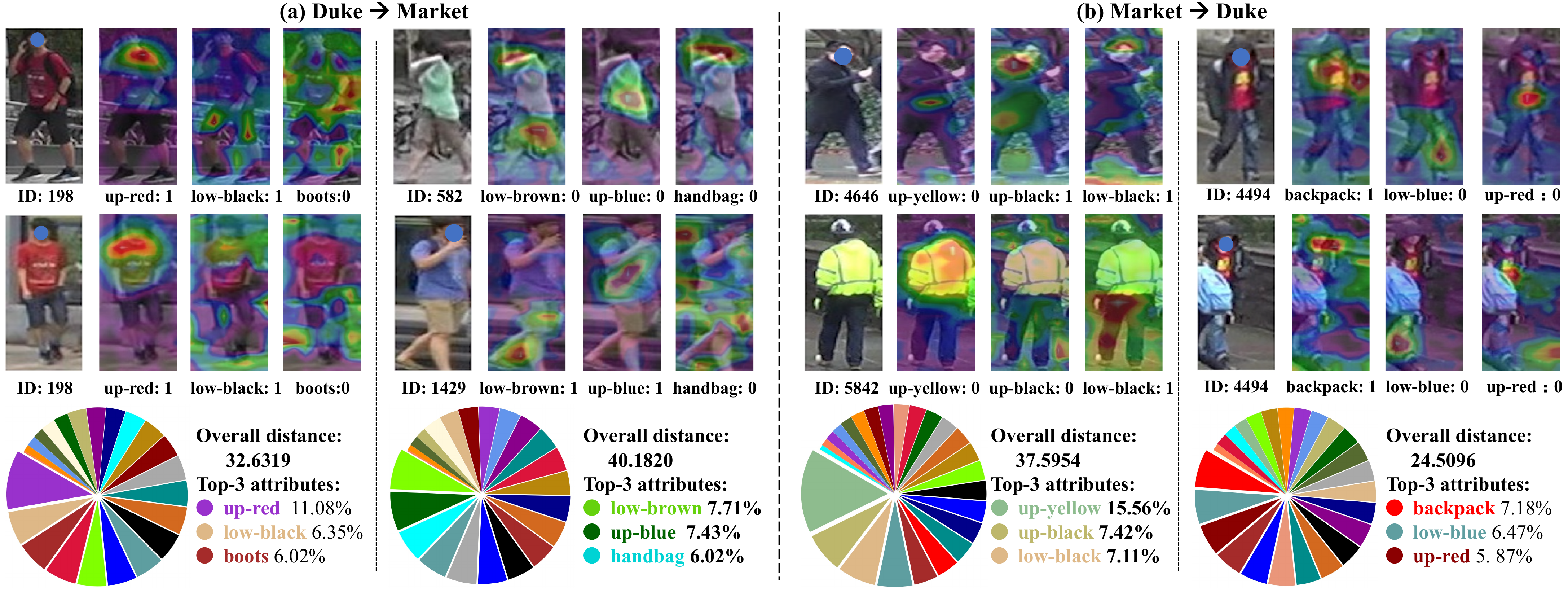}
    \caption{Pairwise examples and explanations for SBS (ResNet-50) under the cross-domain setting: (a) The interpreter is learned on DukeMTMC-ReID and applied to Market-1501 (Duke $\rightarrow$ Market); (b) The interpreter is trained on Market-1501 and tested on DukeMTMC-ReID (Market $\rightarrow$ Duke). For each pair of images, the upper part visualizes the AAMs of the top-3 attributes, which shows that the AAMs are attended to the discriminative attributes. The lower part provides the overall distance and contributions of the top-3 attributes to show the most contributed attributes discovered by the interpreter. (Best viewed in color.)}
\label{fig:figure7} \vspace{0mm}
\end{figure*}

For DukeMTMC-ReID, we select 23 attributes for the interpreter, i.e., gender (female/male), shoe type (boots/other shoes), wearing a hat (yes/no), carrying a backpack (yes/no), carrying handbag(yes/no), carrying other bags (yes/no), the color of shoes (dark/bright), length of upper clothing (long/short), 8 colors of upper clothing (black, white, red, purple, gray, blue, green, and brown) and 7 colors of lower clothing (black, white, red, gray, blue, green, and brown).
The statistics of attributes on DukeMTMC-ReID~\cite{iccv/ZhengZY17} annotated by~\cite{pr/LinZZWHYY19} are shown in Figure~\ref{fig:figure5}.

\subsection{Visualizations of AAMs on Different Datasets}
\label{subsec:result}

In this subsection, we exploit an interpreter learned for the target model, i.e., SBS (ResNet-50)~\cite{corr/abs-2006-02631} to further study the attention maps learned by the interpreter.
Here we generate the average attention maps of individual attributes on the testing set of Market-1501~\cite{iccv/ZhengSTWWT15} and DukeMTMC-ReID~\cite{iccv/ZhengZY17}, as shown in Figure~\ref{fig:figure6} (a) and (b), respectively.
In Figure~\ref{fig:figure6} (a) or (b), the left part shows the positive average attention maps which are obtained from all images that contain a certain attribute.
The right part shows the negative average attention maps obtained from all images that do not contain that attribute.

From the average attention maps, we can observe that:
\textbf{1)} Overall we can find that the interpreter can effectively focus on the salient regions of most attributes, which is consistent with the observation of humans.
\textbf{2)} For the large-area attributes such as the colors of upper and lower clothes, the interpreter can accurately focus on the corresponding regions.
However, there are some worse examples, such as ``low-green''.
This may be because these attributes have very few samples in the dataset, as shown in Figure~\ref{fig:figure5}.
\textbf{3)} For those small but discriminative attributes like bag, hair, and handbag, the interpreter can also attend to the areas where the objects are most likely to appear.
With this observation, we can figure out how the interpreter captures the differences between different attributes through attention. 

Moreover, we have several interesting findings:
\textbf{1)} The salient region of ``gender'' is mainly focused on the head and upper body of a person, which is similar to the attention of humans.
2) For DukeMTMC-ReID, the activation of ``boots'' is focused on both head and feet while the attention of ``shoes'' is only on the feet.
This reflects that the ReID model learns biased knowledge about ``boots'' since the correlation among ``boots'', ``hair'', and ``female'' is very high as discussed in~\cite{pr/LinZZWHYY19}.
Therefore, the interpreter can help researchers and users find the biases in the datasets and improve the ReID models.

\begin{table}[t]
\begin{center}
    \small
    \begin{threeparttable}
    \begin{tabular}{c|c|cc|cc}
    \toprule
    Model                           	& $N$   & Rank-1        & mAP       & X-mAP$_e$ & X-mAP$_c$   \\
    \midrule
    ResNet-50                           &  -    & 94.77         & 87.15     &   -       &   -       \\
    \cmidrule{1-6}
    \multirow{5}{*}{Interpreter} 	    &  1    & 94.76         & 86.54     & 73.79     & 96.53     \\
                                    	&  2    & 93.98         & 86.27     & 74.31     & 96.36     \\
                                    	&  3    & 94.74         & 87.11     & 74.29     & 96.59     \\
                                    	&  4    & 94.52         & 86.91     & 74.47     & 96.79     \\
                                    	&  5    & 94.89         & 87.01     & 74.24     & 96.12    \\
    \bottomrule
    \end{tabular}
    \caption{Results of interpreters sharing different numbers $N$ of stages from the target model. The results show that the interpreters have close performance for distance distillation.}
    \label{tab:table4}
    \end{threeparttable}  \vspace{-7mm}
\end{center}
\end{table}

\subsection{Interpretation for Cross-domain Setting}
\label{subsec:Cross-domain}

In Figure~\ref{fig:figure7}, we show several pairwise images and their explanations for SBS (ResNet-50) models trained on Market-1501 and DukeMTMC-ReID under the cross-domain setting.
For each pair of images, we visualize the attribute-guided attention maps (AAMs) of the top-3 attributes and list the contributions of top-3 attributes to the overall distance.

From the AAMs, we can observe that for the attributes with similar distributions in two datasets, such as ``up-red'' and ``backpack'', the interpreter can effectively focus on corresponding regions under the cross-domain setting and make appropriate explanations.
However, for the unique attributes of a certain dataset, e.g., ``boots'' that only exist in DukeMTMC-ReID, the interpreter would learn nothing about these attributes and generate incorrect explanations. 
To solve this problem, a straightforward strategy is to exploit a more comprehensive dataset with more diverse attributes, such as MTMS17~\cite{cvpr/WeiZ0018}.

\begin{figure}[t]
    \centering
    \includegraphics[width=0.9999\linewidth]{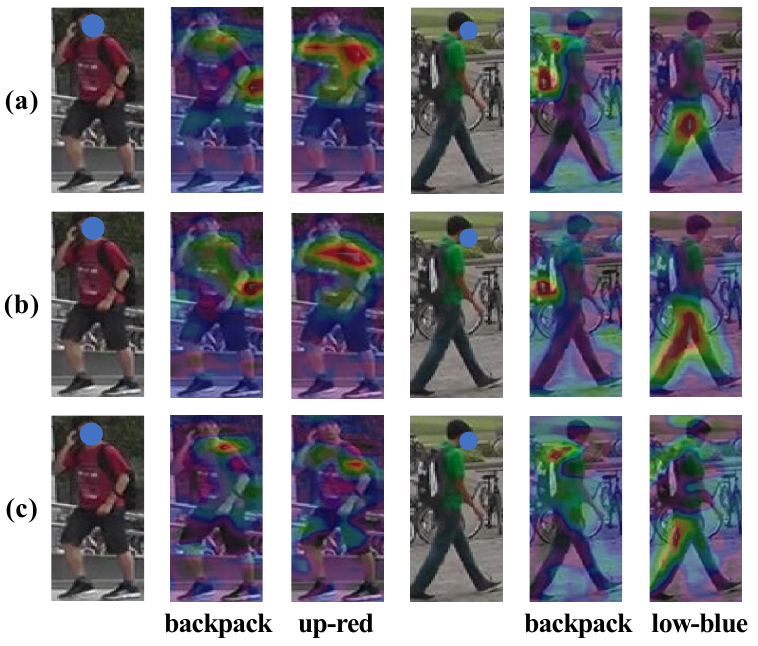}
    \caption{Attention maps from interpreters sharing different stages $N$ from target models. (a) N = 1. (b) N = 3. (c) N = 5. The visualizations reflect the difference of interpreters using different levels of features from the target model. (Best viewed in color.)}
\label{fig:figure8}\vspace{-5mm}
\end{figure}

\subsection{Study on Different Designs of Interpreter}
\label{subsec:explore}
We explore different designs of the interpreter by comparison of interpreter networks that share different numbers of stages from the target ReID model.
Table~\ref{tab:table4} lists the results of interpreters sharing from only one stage to sharing all five stages ($N= 1 \sim 5$).
We visualize several AAMs generated by different interpreters ($N= 1, 3, 5$) in Figure~\ref{fig:figure8}.

From Table~\ref{tab:table4}, we can find that the quantitative metrics of all variants are very close.
This means that all stages of the target ReID model may implicitly learn the knowledge to distinguish attributes.
Only based on this observation, we might conclude that sharing five stages and only training the ADH module is the best choice since it needs to train much fewer parameters.
However, as shown in Figure~\ref{fig:figure8}, the attention maps generated by different interpreters are of great difference.
When $N=1$, the attention maps are more scattered since the receptive field of the lower stage is relatively small which only focuses on local details.
If $N=5$, the attention maps are more focused on the centers of objects because the higher stages learn more high-level semantic concepts.
By sharing parameters from the middle stage, i.e., $N=3$, it can reach an equilibrium point between the low-level patterns and high-level semantics, which is more suitable for the representation of attributes.
Therefore, our interpreter shares three stages from the target models.

\end{appendix}

{\small
\balance
\bibliographystyle{ieee_fullname}
\bibliography{iccv2021}
}

\end{document}